\documentclass[10pt,twocolumn,letterpaper]{article}
\usepackage{cvpr}
\usepackage{times}
\usepackage{epsfig}
\usepackage{graphicx}
\usepackage{amsmath}
\usepackage{amssymb}
\usepackage{multirow}
\usepackage{enumitem}
\usepackage[breaklinks=true,bookmarks=false]{hyperref}
\cvprfinalcopy 
\def\cvprPaperID{1530} 

\begin{document}
\title{Text2Scene: Generating Compositional Scenes from Textual Descriptions}
\author{Fuwen Tan$^1$  \quad Song Feng$^2$ \quad Vicente Ordonez$^1$ \\
  $^1$University of Virginia, $^2$IBM Thomas J. Watson Research Center.\\
  {\tt fuwen.tan@virginia.edu, sfeng@us.ibm.com, vicente@virginia.edu}
}
\maketitle
\begin{abstract}
In this paper, we propose Text2Scene, a model that generates various forms of compositional scene representations from natural language descriptions. Unlike recent works, our method does NOT use Generative Adversarial Networks (GANs). Text2Scene instead learns to sequentially generate objects and their attributes (location, size, appearance, etc) at every time step by attending to different parts of the input text and the current status of the generated scene. We show that under minor modifications, the proposed framework can handle the generation of different forms of scene representations, including cartoon-like scenes, object layouts corresponding to real images, and synthetic images. Our method is not only competitive when compared with state-of-the-art GAN-based methods using automatic metrics and superior based on human judgments but also has the advantage of producing interpretable results.
\end{abstract}

\section{Introduction}
Generating images from textual descriptions has recently become an active research topic~\cite{reed_txt2img,zhang2017stackgan,hdgan,sg2im,attngan,txt2layout2img}. This interest has been partially fueled by the adoption of Generative Adversarial Networks (GANs)~\cite{goodfellow2014generative} which have demonstrated impressive results on a number of image synthesis tasks. 
Synthesizing images from text requires a level of language and visual understanding which could lead to applications in image retrieval through natural language queries, representation learning for text, and automated computer graphics and image editing applications. 

\begin{figure}[t]
\centering
\includegraphics[width=0.48\textwidth]{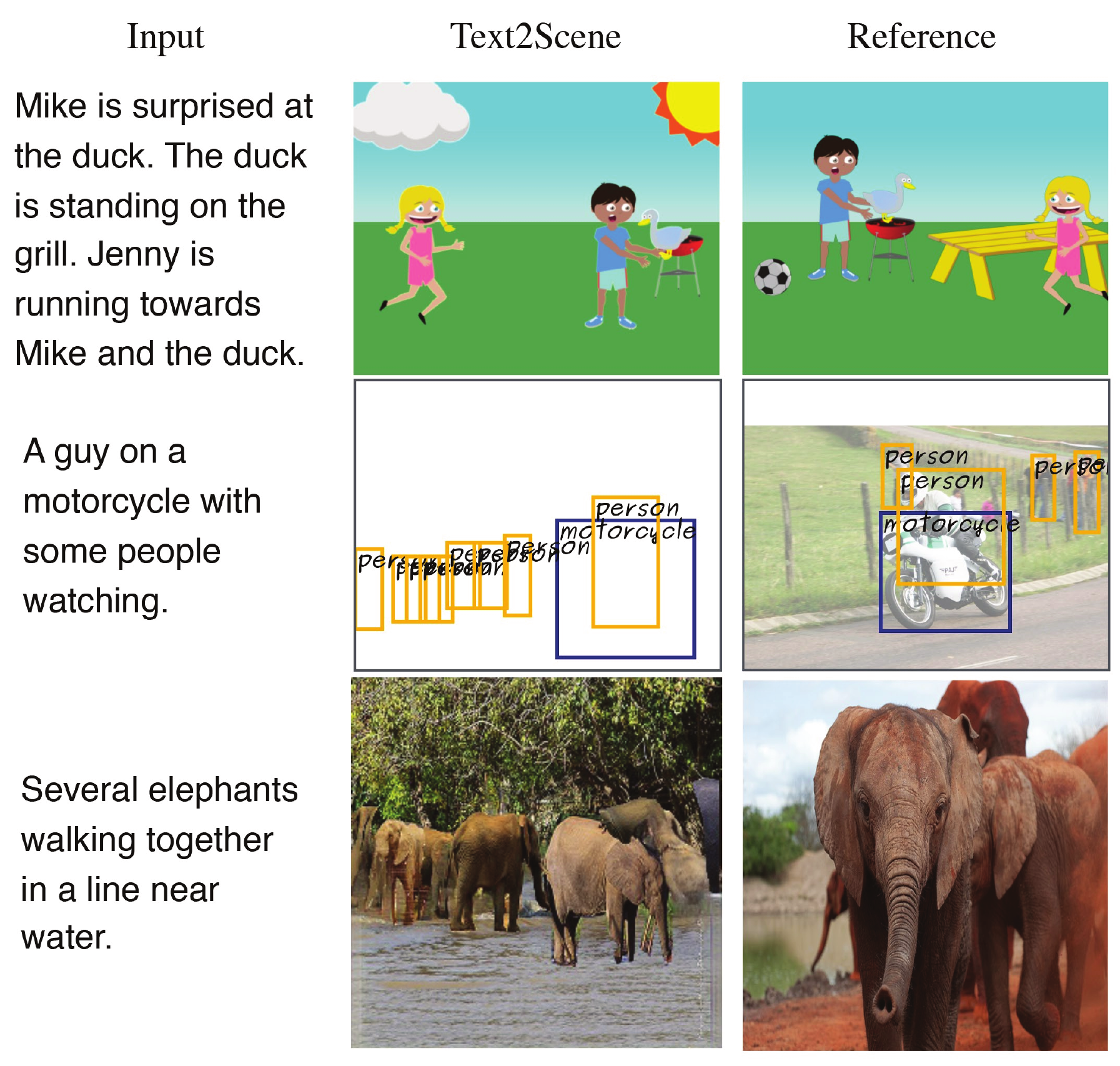}
  \caption{Sample inputs (left) and outputs of our Text2Scene model (middle), along with \emph{ground truth} reference scenes (right) for the generation of abstract scenes (top), object layouts (middle), and synthetic image composites (bottom).}
  \label{fig:teaser}
   \vspace{-0.15in}
\end{figure}

\begin{figure*}[t]
  \centering
  \includegraphics[width=0.85\textwidth]{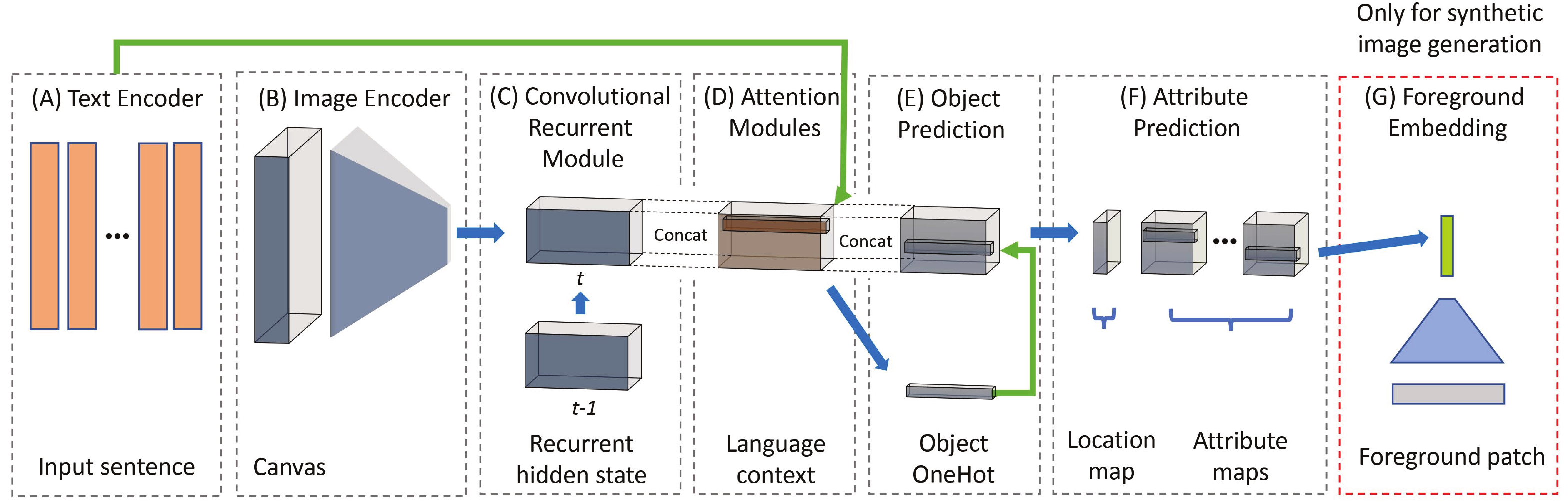}
  \caption{Overview of Text2Scene. Our general framework consists of (A) a Text Encoder that produces a sequential representation of the input, (B) an Image Encoder that encodes the current state of the generated scene, (C) a Convolutional Recurrent Module that tracks, for each spatial location, the history of what have been generated so far, (D-F) two attention-based predictors that sequentially focus on different parts of the input text, first to decide what object to place, then to decide what attributes to be assigned to the object, and (G) an optional foreground embedding step that learns an appearance vector for patch retrieval in the synthetic image generation task.}
  \label{fig:pipeline}
\end{figure*}

In this work, we introduce Text2Scene, a model to interpret visually descriptive language in order to generate compositional scene representations. We specifically focus on generating a scene representation consisting of a list of objects, along with their attributes (e.g. location, size, aspect ratio, pose, appearance). We adapt and train models to generate three types of scenes as shown in Figure~\ref{fig:teaser}, (1) Cartoon-like scenes as depicted in the Abstract Scenes dataset~\cite{Zitnick2013b}, (2) Object layouts corresponding to image scenes from the COCO dataset~\cite{Lin2014}, and (3) Synthetic scenes corresponding to images in the COCO dataset~\cite{Lin2014}. We propose a unified framework to handle these three seemingly different tasks with unique challenges. 
Our method, unlike recent approaches, does not rely on Generative Adversarial Networks (GANs)~\cite{goodfellow2014generative}. Instead, we produce an interpretable model that iteratively generates a scene by predicting and adding new objects at each time step. Our method is superior to the best result reported in Abstract Scenes~\cite{Zitnick2013b}, and provides near state-of-the-art performance on COCO~\cite{Lin2014} under automatic evaluation metrics, and state-of-the-art results when evaluated by humans.

Generating rich textual representations for scene generation is a challenging task. For instance, input textual descriptions could hint only indirectly at the presence of attributes -- e.g.~in the first example in Fig.~\ref{fig:teaser} the input text~``Mike is surprised'' should change the facial attribute on the generated object ``Mike''. Textual descriptions often have complex information about relative spatial configurations -- e.g.~in the first example in Fig.~\ref{fig:teaser} the input text~``Jenny is running towards Mike and the duck'' makes the orientation of ``Jenny'' dependent on the positions of both ``Mike'', and ``duck''. In the last example in Fig.~\ref{fig:teaser} the text ``elephants walking together in a line'' also implies certain overall spatial configuration of the objects in the scene.
\begin{figure*}[t]
  \centering
  \includegraphics[width=0.99\textwidth]{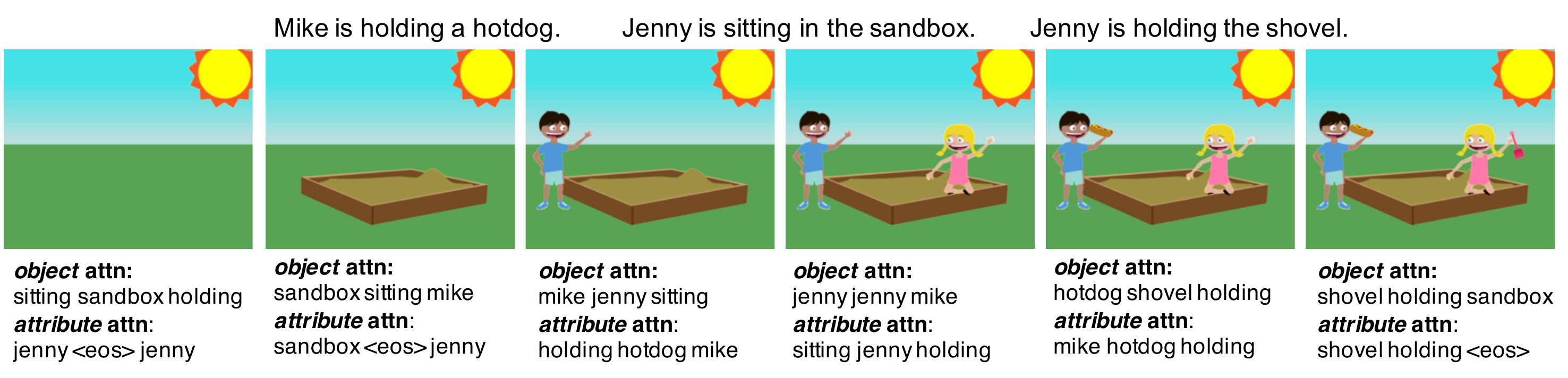}
  \caption{Step-by-step generation of an abstract scene, showing the top-3 attended words for the object prediction and attribute prediction at each time step. Notice how except for predicting the {\it sun} at the first time step, the top-1 attended words in the object decoder are almost one-to-one mappings with the predicted objects. The attended words by the attribute decoder also correspond semantically to useful information for predicting either pose or location, e.g. to predict the location of the {\it hotdog} at the fifth time step, the model attends to {\it mike} and {\it holding}.}
  \label{fig:attn}
\end{figure*}
We model this text-to-scene task using a sequence-to-sequence approach where objects are placed sequentially on an initially empty canvas (see an overview in Fig~\ref{fig:pipeline}). Generally, Text2Scene, consists of a text encoder (Fig~\ref{fig:pipeline} (A)) that maps the input sentence to a set of latent representations, an image encoder (Fig~\ref{fig:pipeline} (B)) which encodes the current generated canvas, a convolutional recurrent module (Fig~\ref{fig:pipeline} (C)), which passes the current state to the next step, attention modules (Fig~\ref{fig:pipeline} (D)) which focus on different parts of the input text, an object decoder (Fig~\ref{fig:pipeline} (E)) that predicts the next object conditioned on the current scene state and attended input text, and an attribute decoder (Fig~\ref{fig:pipeline} (F)) that assigns attributes to the predicted object. To the best of our knowledge, Text2Scene is the first model demonstrating its capacities on both abstract and real images, 
thus opening the possibility for future work on transfer learning across domains. 

Our main contributions can be summarized as follows:
\begin{itemize}[noitemsep,topsep=1pt,parsep=3pt,partopsep=0pt]
    \item We propose Text2Scene, a framework to generate compositional scene representations from natural language descriptions. 
    \item We show that Text2Scene can be used to generate, under minor modifications, different forms of scene representations, including cartoon-like scenes, semantic layouts corresponding to real images, and synthetic image composites.
    \item We conduct extensive experiments on the tasks of abstract image generation for the Abstract Scenes dataset~\cite{Zitnick2013b}, semantic layout and synthetic image generations for the COCO dataset~\cite{Lin2014}.
\end{itemize}
\section{Related Work}
\label{gen_inst}
Most research on visually descriptive language has focused on generating captions from images~\cite{farhadi2010every,mason2014nonparametric,kuznetsova2014treetalk,karpathy2015deep,show_and_tell,show_attend_tell,ordonez2016large,anderson2017guided}. 
Recently, there is work in the opposite direction of text-to-image synthesis~\cite{reed2016learning,reed_txt2img,zhang2017stackgan,sg2im, hdgan, attngan, txt2layout2img}. 
Most of the recent approaches have leveraged conditional Generative Adversarial Networks (GANs). 
While these works have managed to generate results of increasing quality, there are major challenges when attempting to synthesize images for complex scenes with multiple interacting objects without explicitly defining such interactions~\cite{yatskar2016stating}. 
Inspired by \textit{the principle of compositionality}~\cite{compositionality}, our model does not use GANs but produces a scene by sequentially generating objects (e.g. in the forms of clip-arts, bounding boxes, or segmented object patches) containing the semantic elements that compose the scene.

Our work is also related to prior research on using abstract scenes to mirror and analyze complex situations in the real world~\cite{Zitnick2013a,Zitnick2013b,fouhey2014predicting,vedantam2015learning}. 
In~\cite{Zitnick2013b}, a graphical model was introduced to generate an abstract scene from textual descriptions. 
Unlike this previous work, our method does not use a semantic parser but is trained end-to-end from input sentences. 
Our work is also related to recent research on generating images from pixel-wise semantic labels~\cite{pix2pix2017, CRN, Semiparametric},  especially~\cite{Semiparametric} which proposed a retrieval-based semi-parametric method for image synthesis given the spatial semantic map. 
Our synthetic image generation model optionally uses the cascaded refinement module in~\cite{Semiparametric} as a post-processing step. 
Unlike these works, our method is not given the spatial layout of the objects in the scene but learns to predict a layout indirectly from text. 

Most closely related to our approach are~\cite{sg2im, imaginethis, txt2layout2img}, and~\cite{kim2017codraw}, as these works also attempt to predict explicit 2D layout representations.
Johnson~et~al~\cite{sg2im} proposed a graph-convolutional model to generate images from structured scene graphs.
The presented objects and their relationships were provided as inputs in the scene graphs, while in our work, the presence of objects is inferred from text. Hong~et~al~\cite{txt2layout2img} targeted image synthesis using conditional GANs but unlike prior works, it generated layouts as intermediate representations in a separably trained module. 
Our work also attempts to predict layouts for photographic image synthesis but unlike~\cite{txt2layout2img}, we generate pixel-level outputs using a semi-parametric retrieval module without adversarial training and demonstrate superior results.
Kim~et~al~\cite{kim2017codraw} performed pictorial generation from chat logs, while our work uses text which is considerably more underspecified. Gupta~et~al~\cite{imaginethis} proposed a semi-parametric method to generate cartoon-like pictures. However the presented objects were also provided as inputs to the model, and the predictions of layouts, foregrounds and backgrounds were performed by separably trained modules. 
Our method is trained end-to-end and goes beyond cartoon-like scenes. 
To the best of our knowledge, our model is the first work targeting various types of scenes (e.g. abstract scenes, semantic layouts and composite images) under a unified framework.
\section{Model}
\label{model}
Text2Scene adopts a Seq-to-Seq framework~\cite{seq2seq} and introduces key designs for spatial and sequential reasoning.
Specifically, at each time step, the model modifies a background canvas in three steps:
(1) the model attends to the input text to decide what is the next object to add, or decide whether the generation should end; 
(2) if the decision is to add a new object, the model \textit{zooms in} the language context of the object to \textit{decide} its attributes (e.g.~pose, size) and relations with its surroundings (e.g.~location, interactions with other objects);
(3) the model refers back to the canvas and \textit{grounds} (places) the extracted textual attributes into their corresponding visual representations.

To model this procedure, Text2Scene consists of a text encoder, which takes as input a sequence of $M$ words $w_i$ (section~\ref{sec:text-encoder}), an object decoder, which predicts sequentially $T$ objects $o_t$, and an attribute decoder that predicts for each $o_t$ their locations $l_t$ and a set of $k$ attributes $\{R^k_t\}$ (section~\ref{sec:object-decoders}). The scene generation starts from an initially empty canvas $B_0$ that is updated at each time step.
In the synthetic image generation task, we also jointly train a foreground patch embedding network (section~\ref{sec:embedding}) and treat the embedded vector as a target attribute. Figure~\ref{fig:attn} shows a step-by-step generation of an abstract scene.

\subsection{Text Encoder} 
\label{sec:text-encoder}
Our text encoder is a bidirectional recurrent network with Gated Recurrent Units (GRUs). 
For a given sentence, we compute for each word $w_i$:
\begin{equation} 
    h_i^E = \text{BiGRU}(x_i, h_{i-1}^E, h_{i+1}^E),
\end{equation}
Here BiGRU is a bidirectional GRU cell,
$x_i$ is a word embedding vector corresponding to the i-th word $w_i$, 
and $h_i^E$ is a hidden vector encoding the current word and its context. 
We use the pairs $[h_i^E; x_i]$, the concatenation of $h_i^E$ and $x_i$, as the encoded text feature.
\vspace{0.01in}
\subsection{Object and Attribute Decoders}
\label{sec:object-decoders}
At each step $t$, our model predicts the next object $o_t$ from an object vocabulary $\mathcal{V}$ and its $k$ attributes $\{R^k_t\}$, using text feature $\{[h_i^E; x_i]\}$ and the current canvas $B_t$ as input.
For this part, we use a convolutional network (CNN) $\Omega$ to encode $B_t$ into a $\mathcal{C} \times H \times W$ feature map, representing the current scene state.
We model the history of the scene states \{$h_t^D$\} with a convolutional GRU (ConvGRU):
\begin{equation} 
    \label{eq:encoder}
    h_t^D = \text{ConvGRU}(\Omega(B_{t}), h_{t-1}^D),
\end{equation}
The initial hidden state is created by spatially replicating the last hidden state of the text encoder. 
Here $h_t^D$ provides an informative representation of the temporal dynamics of each spatial (grid) location in the scene. 
Since this representation might fail to capture small objects, a one-hot vector of the object predicted at the previous step $o_{t-1}$ is also provided as input to the downstream decoders.
The initial object is set as a special start-of-scene token.

\vspace{0.01in}
\noindent{\bf Attention-based Object Decoder:} Our object decoder is an attention-based model that outputs the likelihood scores of all possible objects in an object vocabulary $\mathcal{V}$. 
It takes as input the recurrent scene state $h_{t}^D$, text features $\{[h_{i}^E; x_i]\}$ and the previously predicted object $o_{t-1}$:
\begin{gather}
    \label{eq:object-visual-attention}
    u^o_t = \text{AvgPooling}(\Psi^o(h_t^D)),\\
    \label{eq:object-textual-attention}
    c^o_t = \Phi^o([u^o_t;o_{t-1}], \{[h_{i}^E; x_i]\}),\\
    \label{eq:object-classifier}
    p(o_t) \propto \Theta^o([u^o_t;o_{t-1};c^o_{t}]),
\end{gather}
here $\Psi^o$ is a convolutional network with spatial attention on $h_t^D$, similar as in~\cite{show_attend_tell}. 
The goal of $\Psi^o$ is to collect the spatial contexts necessary for the object prediction, e.g.~what objects have already been added.
The attended spatial features are then fused into a vector $u^o_t$ by average pooling. 
$\Phi^o$ is the text-based attention module, similar as in~\cite{attention}, which uses $[u^o_t;o_{t-1}]$ to attend to the language context $\{[h_{i}^E; x_i]\}$ and collect the context vector $c^o_t$.
Ideally, $c^o_t$ encodes information about all the described objects that have not been added to the scene thus far.
$\Theta^o$ is a two-layer perceptron predicting the likelihood of the next object $p(o_t)$ from the concatenation of $u^o_t$, $o_{t-1}$, and $c^o_{t}$, using a softmax function.

\vspace{0.01in}
\noindent{\bf Attention-based Attribute Decoder}
The attribute set corresponding to the object $o_t$ can be predicted similarly. 
We use another attention module $\Phi^a$ to ``zoom in'' the language context of $o_t$, extracting a new context vector $c^a_t$.
Compared with $c^o_t$ which may contain information of all the objects that have not been added yet, $c^a_t$ focuses more specifically on contents related to the current object $o_t$. For each spatial location in $h_t^D$, the model predicts a location likelihood $l_t$, and a set of attribute likelihoods $\{R^k_t\}$.
Here, possible locations are discretized into the same spatial resolution of $h_t^D$. In summary, we have:
\begin{gather}
    c^a_t = \Phi^a(o_{t}, \{[h_{i}^E; x_i]\})\\
    u^a_t = \Psi^a([h_t^D; c^a_t]) \\
    \label{eq:attribute-classifier}
    p(l_t, \{R^k_t\}) = \Theta^a([u^a_t;o_t;c^a_t]),
\end{gather}
$\Phi^a$ is a text-based attention module aligning $o_t$ with the language context $\{[h_{i}^E; x_i]\}$.
$\Psi^a$ is an image-based attention module aiming to find an affordable location to add $o_t$.
Here $c^a_t$ is spatially replicated before concatenating with $h_t^D$.
The final likelihood map $p(l_t, \{R^k_t\})$ is predicted by a convolutional network $\Theta^a$, followed by softmax classifiers for $l_t$ and discrete $\{R^k_t\}$.
For continuous attributes $\{R^k_t\}$ such as the appearance vector $Q_t$ for patch retrieval (next section), we normalize the output using an $\ell_2$-norm. 

\subsection{Foreground Patch Embedding}
\label{sec:embedding}
We predict a particular attribute: an appearance vector $Q_t$, only for the model trained to generate synthetic image composites (i.e. images composed of patches retrieved from other images). 
As with other attributes, $Q_t$ is predicted for every location in the output feature map which is used at test time to retrieve similar patches from a pre-computed collection of object segments from other images. 
We train a patch embedding network using a CNN which reduces the foreground patch in the target image into a 1D vector $F_t$. 
The goal is to minimize the $\ell_2$-distance between $Q_t$ and $F_t$ using a triplet embedding loss~\cite{facenet} that encourages a small distance $||Q_t, F_t||_{2}$ but a larger distance with other patches $||Q_t, F_k||_{2}$. 
Here $F_k$ is the feature of a "negative" patch, which is randomly selected from the same category of $F_t$: 
\begin{equation}
    L_{triplet}(Q_t, F_t) = \\
    max\{||Q_t, F_t||_2 - ||Q_t, F_k||_2 + \alpha, 0\} 
\end{equation}
$\alpha$ is a margin hyper-parameter.

\begin{table*}[h]
  \centering
  \begin{tabular}{lllllllll}
    \hline 
    \multirow{2}{*}{Methods} & \multicolumn{2}{c}{U-obj} & \multicolumn{2}{c}{B-obj} & \multirow{2}{*}{Pose} & \multirow{2}{*}{Expr} & U-obj & B-obj\\
       & Prec & Recall & Prec & Recall &  &  & Coord & Coord\\
    \hline 
    Zitnick et al.~\cite{Zitnick2013b} & 0.722 & 0.655 & 0.280 & 0.265 & 0.407 & 0.370 & \textbf{0.449} & 0.416   \\
    Text2Scene (w/o attention) & 0.665 & 0.605 & 0.228 & 0.186 & 0.305 & 0.323 & 0.395 & 0.338 \\
    Text2Scene (w object attention) & 0.731 & 0.671 & 0.312 & 0.261 & 0.365 & 0.368 & 0.406 & 0.427 \\
    Text2Scene (w both attentions) & 0.749 & 0.685 & 0.327 & 0.272 & 0.408 & 0.374 & 0.402 & 0.467 \\
    Text2Scene (full) & \textbf{0.760} & \textbf{0.698} & \textbf{0.348} & \textbf{0.301} & \textbf{0.418} & \textbf{0.375} & 0.409 & \textbf{0.483} \\
    \hline
  \end{tabular}
  \vspace{0.05in}
  \caption{Quantitative evaluation on the Abstract Scenes dataset. Our full model performs better in all metrics except U-obj Coord which measures exact object coordinates. It also shows that our sequential attention approach is effective.}
  \label{quan_abstract}
\end{table*}

\begin{table*}[t]
  \centering
  \begin{tabular}{ll | ll | cccc}
    \hline 
    \multirow{2}{*}{Methods} & \multirow{2}{*}{Scores} & \multirow{2}{*}{\(\geq 1\)} & \multirow{2}{*}{\(\geq2\)} & Obj-Single & Obj-Pair & Location & Expression \\
     &  &  &   & {\scriptsize \texttt{sub-pred}} & {\scriptsize\texttt{sub-pred-obj}} & {\scriptsize\texttt{pred:loc}} & {\scriptsize\texttt{pred:expr}}\\
    \hline 
    Reference & 0.919 & 1.0 & 0.97  
    & 
    0.905 & 0.88 & 0.933 & 0.875 \\
    \hline
    Zitnick et al.~\cite{Zitnick2013b} & 0.555  & 0.92 & 0.49 
    & 
    0.53 & 0.44 & \textbf{0.667} & 0.625 \\
    Text2Scene (w/o attention) & 0.455 & 0.75 & 0.42
    & 
    0.431 & 0.36 & 0.6 & 0.583 \\
    Text2Scene (full) & \textbf{0.644}  & \textbf{0.94} & \textbf{0.62}
    & 
    \textbf{0.628} & \textbf{0.48} & \textbf{0.667} & \textbf{0.708} \\
    \hline
  \end{tabular}
    \vspace{0.05in}
  \caption{Human evaluation on Abstract Scenes. Each scene is generated from three textual statements. Users are asked to rate if the generated scene validates each input statement. Our method generates scenes that abide by at least one of the statements 94\% of the times, and by at least two statements 64\%, and is superior in all types of statements except Location.}
  \label{human_abstract}
\end{table*}

\begin{table*}[h]
  \centering
  \begin{tabular}{lllllllll}
    \hline 
    Methods & B1 & B2 & B3 & B4 & METEOR & ROUGE & CIDEr & SPICE\\
    \hline
    Captioning from True Layout~\cite{obj2textEMNLP2017}  & 0.678 & 0.492 & 0.348 & 0.248 & 0.227 & 0.495 & 0.838 & 0.160 \\
    \hline
    Text2Scene (w/o attention) & 0.591 & 0.391 & 0.254 & 0.169 & 0.179 & 0.430 & 0.531 & 0.110 \\
    Text2Scene (w object attention)  & 0.591 & 0.391 & 0.256 & 0.171 & 0.179 & 0.430 & 0.524 & 0.110 \\
    Text2Scene (w both attentions)  & 0.600 & 0.401 & 0.263 & 0.175 & 0.182 & 0.436 & 0.555 & 0.114 \\
    Text2Scene (full)  & \textbf{0.615} & \textbf{0.415} & \textbf{0.275} & \textbf{0.185} & \textbf{0.189} & \textbf{0.446} & \textbf{0.601} & \textbf{0.123}\\
    \hline
  \end{tabular}
  \vspace{0.05in}
  \caption{Quantitative evaluation on the layout generation task. Our full model generates more accurate captions from the generated layouts than the baselines. We also include caption generation results from ground truth layouts as an upper bound on this task.}
  \label{quan_coco}
\end{table*}

\subsection{Objective}
\label{sec:objective}
The loss function for a given example in our model with reference values $(o_t, l_t, \{R^k_t\}, F_t)$ is:
\begin{align*}
    L = & -w_o\sum_t \text{log } p(o_{t}) -w_l\sum_t \text{log }p(l_{t}) \\
    &-\sum_kw_k\sum_t \text{log } p({R^k_{t}}) + w_e\sum_t L_{triplet}(Q_t, F_t)\\
    &+w_a^O L^{O}_{attn} + w_a^A L^{A}_{attn}, 
\end{align*}
where the first three terms are negative log-likelihood losses corresponding to the object, location, and discrete attribute softmax classifiers.
$L_{triplet}$ is a triplet embedding loss optionally used for the synthetic image generation task.
$L^{*}_{attn}$ are regularization terms inspired by the doubly stochastic attention module proposed in~\cite{show_attend_tell}.
Here $L^{*}_{attn}=\sum_i [1-\sum_t\alpha^{*}_{ti}]^2$ where $\{\alpha^o_{ti}\}$ and $\{\alpha^a_{ti}\}$ are the attention weights from $\Phi^o$ and $\Phi^a$ respectively. 
These regularization terms encourage the model to distribute the attention across all the words in the input sentence so that it will not miss any described objects.
Finally, $w_o$, $w_l$, $\{w_k\}$, $w_e$, $w_a^O$, and $w_a^A$ are hyperparameters controlling the relative contribution of each loss.

\vspace{0.01in}
\noindent{\bf Details for different scene generation tasks}
In the Abstract Scenes generation task, $B_t$ is represented directly as an RGB image. 
In the layout generation task, $B_t$ is a 3D tensor with a shape of ($\mathcal{V}, H, W$), where each spatial location contains a one-hot vector indicating the semantic label of the object at that location.
Similarly, in the synthetic image generation task, $B_t$ is a 3D tensor with a shape of (3$\mathcal{V}, H, W$), where every three channels of this tensor encode the color patches of a specific category from the background canvas image.
For the foreground embedding module, we adopt the canvas representation in~\cite{Semiparametric} to encode the foreground patch for simplicity.
As the composite images may exhibit gaps between patches, we also leverage the stitching network in~\cite{Semiparametric} for post-processing.
Note that the missing region may also be filled by any other inpainting approaches. 
Full details about the implementation of our model can be found in the supplementary material. Our code and data are publicly available\footnote{\url{https://github.com/uvavision/Text2Scene}}.

\begin{figure}[h]
  \centering
  \includegraphics[width=0.46\textwidth]{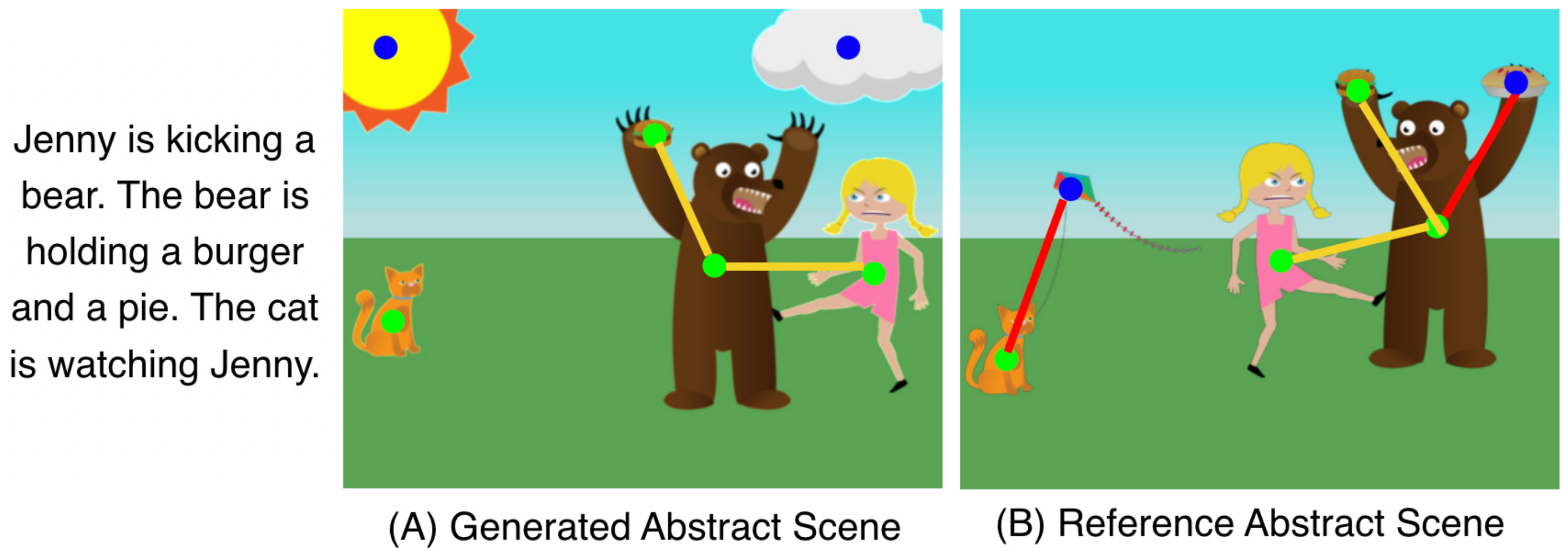}
  \caption{Evaluation metrics for the abstract scene generation task (best viewed in color): the green dots show the common \texttt{U-obj} between the reference (B) and the generated abstract scene (A), the blue dots show the missing and mispredicted objects. Similarly, the yellow lines show the common \texttt{B-obj} and the red lines show the missing and mispredicted \texttt{B-obj}. 
  The \texttt{U-obj} precision/recall for this example is 0.667/0.667, the \texttt{B-obj} precision/recall is 1.0/0.5.}
  \label{fig:metric}
\end{figure}

\section{Experiments}
We conduct experiments on three text-to-scene tasks: 
(I) constructing abstract scenes of clip-arts in the Abstract Scenes~\cite{Zitnick2013b} dataset; 
(II) predicting semantic object layouts of real images in the COCO~\cite{Lin2014} dataset; and
(III) generating synthetic image composites in the COCO~\cite{Lin2014} dataset.

\begin{table*}[t]
  \centering
  \begin{tabular}{lllllllllll}
    \hline 
    Methods & IS & B1 & B2 & B3 & B4 & METEOR & ROUGE & CIDEr & SPICE\\
    \hline
    Real image & 36.00$\pm$0.7 &0.730 & 0.563 & 0.428 & 0.327 & 0.262 & 0.545 & 1.012 & 0.188  \\
    \hline
    GAN-INT-CLS~\cite{reed_txt2img}  & 7.88$\pm$0.07 & 0.470 & 0.253 & 0.136 & 0.077 & 0.122 & -- & 0.160 & --  \\
    SG2IM*~\cite{sg2im} & 6.7$\pm$0.1 & 0.504 & 0.294 & 0.178 & 0.116 & 0.141 & 0.373 & 0.289 & 0.070  \\
    StackGAN~\cite{zhang2017stackgan}  & 10.62$\pm$0.19 & 0.486 & 0.278 & 0.166 & 0.106 & 0.130 & 0.360 & 0.216 & 0.057  \\
    HDGAN~\cite{hdgan}  & 11.86$\pm$0.18 & 0.489 & 0.284 & 0.173 & 0.112 & 0.132 & 0.363 & 0.225 & 0.060  \\
    Hong et al~\cite{txt2layout2img}  & 11.46$\pm$0.09 & 0.541 & 0.332 & 0.199 & 0.122 & 0.154 & -- & 0.367 & --  \\
    AttnGan~\cite{attngan} & \textbf{25.89$\pm$0.47} & \textbf{0.640} & \textbf{0.455} & \textbf{0.324} & \textbf{0.235} & \textbf{0.213} & \textbf{0.474} & \textbf{0.693} & \textbf{0.141}  \\
    \hline
    Text2Scene (w/o inpaint.)  & 22.33$\pm$1.58 &0.602 & 0.412 & 0.288 & 0.207 & 0.196 & 0.448 & 0.624 & 0.126  \\
    Text2Scene (w inpaint.)  & 24.77$\pm$1.59 &0.614 & 0.426 & 0.300 & 0.218 & 0.201 & 0.457 & 0.656 & 0.130  \\
    \hline
  \end{tabular}
  \vspace{0.05in}
  \caption{Quantitative evaluation on the synthetic image generation task. Our model is superior on automated metrics than all competing approaches except AttnGan, even without post-processing. *The result of SG2IM is evaluated on the validation set defined in~\cite{sg2im}, which is a subset of the COCO val2014 split.}
  \label{caption_coco}
\end{table*}

\begin{table}[t]
  \centering
  \begin{tabular}{ll}
    \hline 
    & Ratio\\
    \hline
    Text2Scene $>$ SG2IM~\cite{sg2im} & 0.7672 \\
    Text2Scene $>$ HDGAN~\cite{hdgan} & 0.8692 \\
    Text2Scene $>$ AttnGAN~\cite{attngan}& 0.7588\\
    \hline
  \end{tabular}
  \vspace{0.05in}
  \caption{Two-alternative forced-choiced evaluation on the synthetic image generation task against competing methods.  }
  \label{human_eval}
\end{table}

\vspace{0.03in}
\noindent{\bf Task (I): Clip-art Generation on Abstract Scenes}
\label{sec:abstract-scenes}
We use the dataset introduced by \cite{Zitnick2013b}, which contains over 1,000 sets of 10 semantically similar scenes of children playing outside. 
The scenes are composed with 58 clip-art objects.
The attributes we consider for each clip-art object are the location, size ($|R^{size}|=3$), and the direction the object is facing ($|R^{direction}|=2$).
For the person objects, we also explicitly model the pose ($|R^{pose}|=7$) and expression ($|R^{expression}|=5$).
There are three sentences describing different aspects of a scene.
After filtering empty scenes, we obtain $9997$ samples.
Following \cite{Zitnick2013b}, we reserve 1000 samples as the test set and $497$ samples for validation.

\vspace{0.03in}
\noindent{\bf Task (II): Semantic Layout Generation on COCO}
The semantic layouts contain bounding boxes of the objects from 80 object categories defined in the COCO~\cite{Lin2014} dataset. 
We use the val2017 split as our test set and use $5000$ samples from the train2017 split for validation. 
We normalize the bounding boxes and order the objects from bottom to top as the y-coordinates typically indicate the distances between the objects and the camera.
We further order the objects with the same y-coordinate based on their x-coordinates (from left to right) and categorical indices.
The attributes we consider are location, size ($|R^{size}|=17$), and aspect ratio ($|R^{aspect\_ratio}|=17$). 
For the size attribute, we discretize the normalized size range evenly into 17 scales.
We also use 17 aspect ratio scales, which are $\{\frac{1}{9}, \frac{1}{8}, \frac{1}{7}, \frac{1}{6}, \frac{1}{5}, \frac{1}{4}, \frac{1}{3}, \frac{1}{2}, \frac{1}{1}, \frac{2}{1}, \frac{3}{1}, \frac{4}{1}, \frac{5}{1}, \frac{6}{1}, \frac{7}{1}, \frac{8}{1}, \frac{9}{1}\}$.

\vspace{0.03in}
\noindent{\bf Task (III): Synthetic Image Generation on COCO}
We demonstrate our approach by generating synthetic image composites given input captions in COCO~\cite{Lin2014}.
For fair comparisons with alternative approaches, we use the val2014 split as our test set and use $5000$ samples from the train2014 split for validation.
We collect segmented object and stuff patches from the training split. 
The stuff segments are extracted from the training images by taking connected components in corresponding semantic label maps from the COCO-Stuff annotations~\cite{cocostuff}. 
For object segments, we use all 80 categories defined in COCO.
For stuff segments, we use the 15 super-categories defined in~\cite{cocostuff} as the class labels, which results in 95 categories in total.
We order the patches as in the layout generation task but when composing the patches, we always render the object patches in front of the stuff patches.
In our experiment, $Q_t$ and $F_t$ have a dimension of 128. 

\subsection{Evaluation}
\paragraph{Automatic Metrics}

Our tasks pose new challenges on evaluating the models as (1) the three types of scene representations are quite different from each other; and (2) there is no absolute one-to-one correspondence between a sentence and a scene. For the abstract scene generation task, we draw inspiration from the evaluation metrics applied in machine translation~\cite{meteor} but we aim at aligning multimodal visual-linguistic data instead. 
To this end, we propose to compute the following metrics: precision/recall on single objects (\texttt{U-obj}), ``bigram'' object pairs (\texttt{B-obj}); classification accuracies for poses, expressions; Euclidean distances (defined as a Gaussian function with a kernel size of 0.2) for normalized coordinates of \texttt{U-obj} and \texttt{B-obj}. A ``bigram'' object pair is defined as a pair of objects with overlapping bounding boxes as illustrated in Figure \ref{fig:metric}.

In the layout generation experiment, it is harder to define evaluation metrics given the complexity of real world object layouts. 
Inspired by~\cite{txt2layout2img}, we employ caption generation as an extrinsic evaluation.
We generate captions from the semantic layouts using~\cite{obj2textEMNLP2017} and compare them back to the original captions used to generate the scenes. 
We use commonly used metrics for captioning such as BLEU \cite{papineni2002bleu}, METEOR~\cite{meteor}, ROUGE\_L \cite{lin2004rouge}, CIDEr \cite{vedantam2015cider} and SPICE \cite{anderson2016spice}.

For synthetic image generation, we adopt the Inception Score (IS) metric~\cite{inceptionscore} which is commonly used in recent text to image generation methods. However, as IS does not evaluate correspondence between images and captions, we also employ an extrinsic evaluation using image captioning using the Show-and-Tell caption generator~\cite{show_and_tell}  as in~\cite{txt2layout2img}. 

\vspace{0.01in}
\noindent{\bf Baselines}
For abstract scene generation, we run comparisons with~\cite{Zitnick2013b}. 
We also consider variants of our full model: 
(1) Text2Scene (w/o attention): a model without any attention module. 
In particular, we replace Eq.~\ref{eq:object-visual-attention} with a pure average pooling operation on $h^D_t$, discard $c^o_t$ in Eq.~\ref{eq:object-classifier}, discard $c^a_t$ and replace $u^a_t$ with $h^D_t$ in Eq.~\ref{eq:attribute-classifier}. 
(2) Text2Scene (w object attention): a model with attention modules for object prediction but no dedicated attention for attribute prediction.
Specifically, we replace ($u^a_t$, $c^a_t$) with ($h^D_t$, $c^o_t$) in Eq.~\ref{eq:attribute-classifier}.
(3) Text2Scene (w both attentions): a model with dedicated attention modules for both object and attribute predictions but no regularization. 

\begin{figure}[t]
  \centering
  \includegraphics[width=0.48\textwidth]{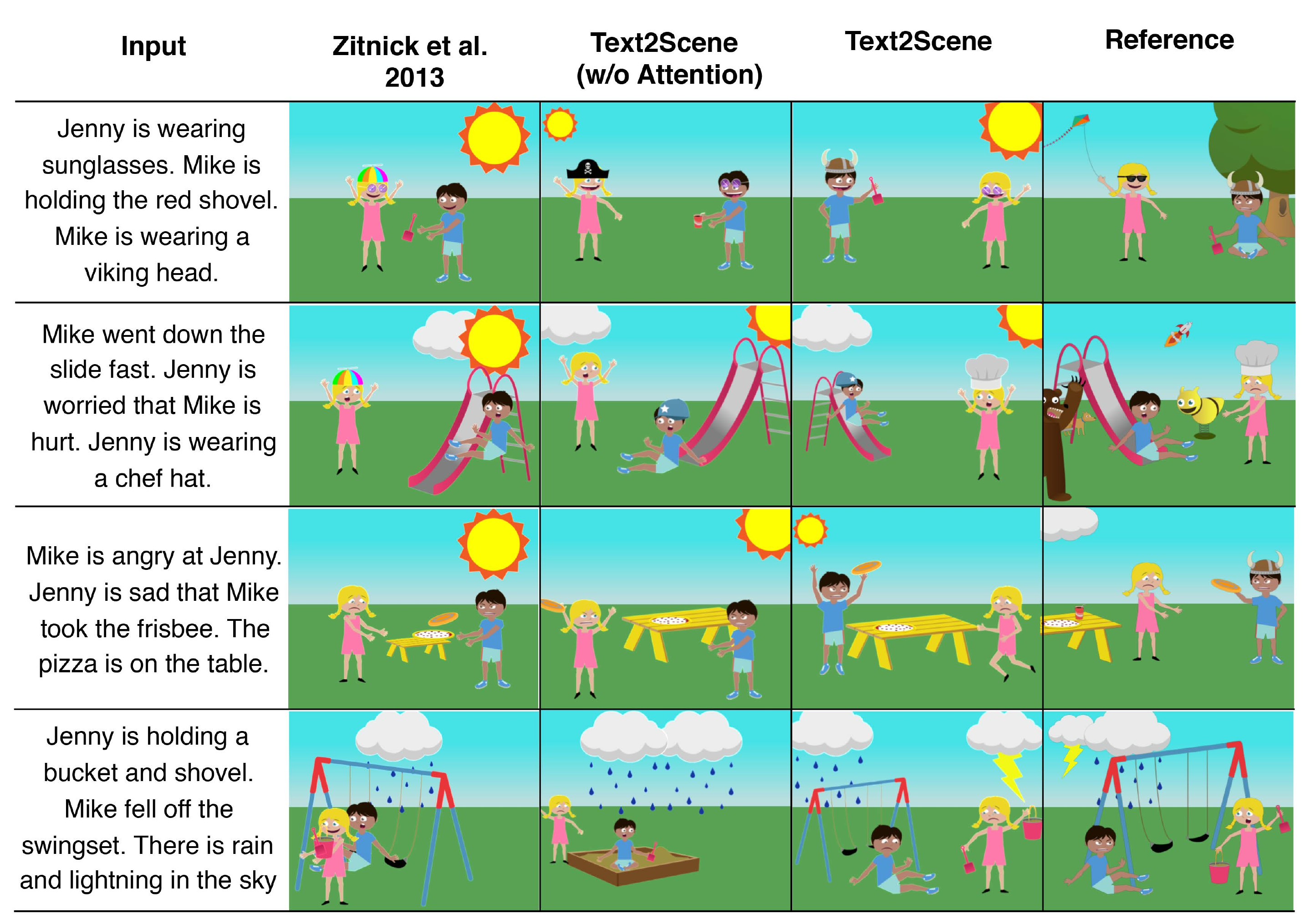}
  \caption{Examples of generated abstract scenes. The first column shows the input text, and the last column shows the reference \emph{true} scene from the dataset.}
  \label{fig:abstract_qual}
\end{figure}

\vspace{0.02in}
\noindent{\bf Human Evaluations}
 We also conduct human evaluations using crowdsourcing on 100 groups of clip-art scenes generated for the Abstract Scene dataset using random captions from the test split. Human annotators are asked to determine whether an input text is a true statement given the generated scene (entailment). 
Each scene in this dataset is associated with three sentences that are used as the statements. 
Each sentence-scene pair is reviewed by three annotators to determine if the entailment is \texttt{true}, \texttt{false} or \texttt{uncertain}.
Ignoring \texttt{uncertain} responses, we use the ratio of the sentence-scene pairs marked as \texttt{true} for evaluation.

We also perform predicate-argument semantic frame analysis~\cite{carreras2005introduction} on our results. Using the semantic parser from~\cite{Zitnick2013b}, we subdivide the sentences as: \texttt{sub-pred} corresponding to sentences referring to only one object, \texttt{sub-pred-obj} corresponding to sentences referring to object pairs with semantic relations, \texttt{pred:loc} corresponding to sentences referring to locations, and \texttt{pred:pa} corresponding to sentences mentioning facial expressions.

For synthetic image generation we use a similar human evaluation as in~\cite{Semiparametric}.
We compare our method against SG2IM~\cite{sg2im}, HDGAN~\cite{hdgan} and AttnGAN~\cite{attngan}.
We resize our generated images to the same resolutions as in these alternative methods, 64 $\times$ 64 for SG2IM~\cite{sg2im}, 256 $\times$ 256 for HDGAN~\cite{hdgan} and AttnGAN~\cite{attngan}.
For each sentence randomly selected from the test set, we present images generated by our method and a competing method and allow the user to choose the one which better represents the text. 
We collect results for 500 sentences. For each sentence, we collect responses from 5 different annotators.

\begin{figure}[t]
  \centering
  \includegraphics[width=0.47\textwidth]{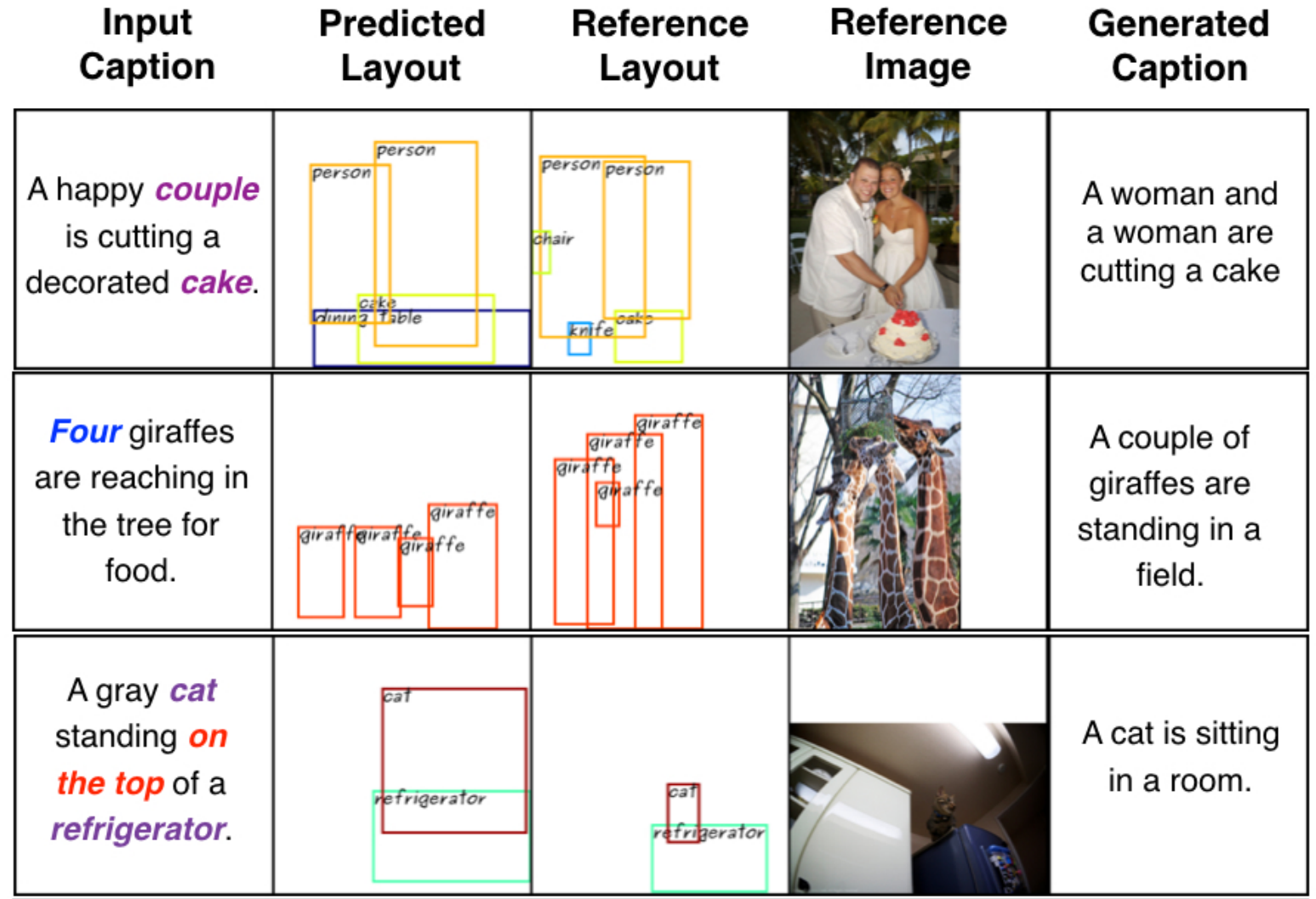}
  \caption{Generated layouts from input captions and generated captions from the predicted layouts (best viewed in color). Our model successfully predicts the presence (purple text) and number of objects (blue text), and their spatial relations (red text).}
  \label{fig:coco_qual}
\end{figure}

\subsection{Results and Discussion}
\label{results}

\begin{figure*}[t]
  \centering
  \includegraphics[width=0.8\textwidth]{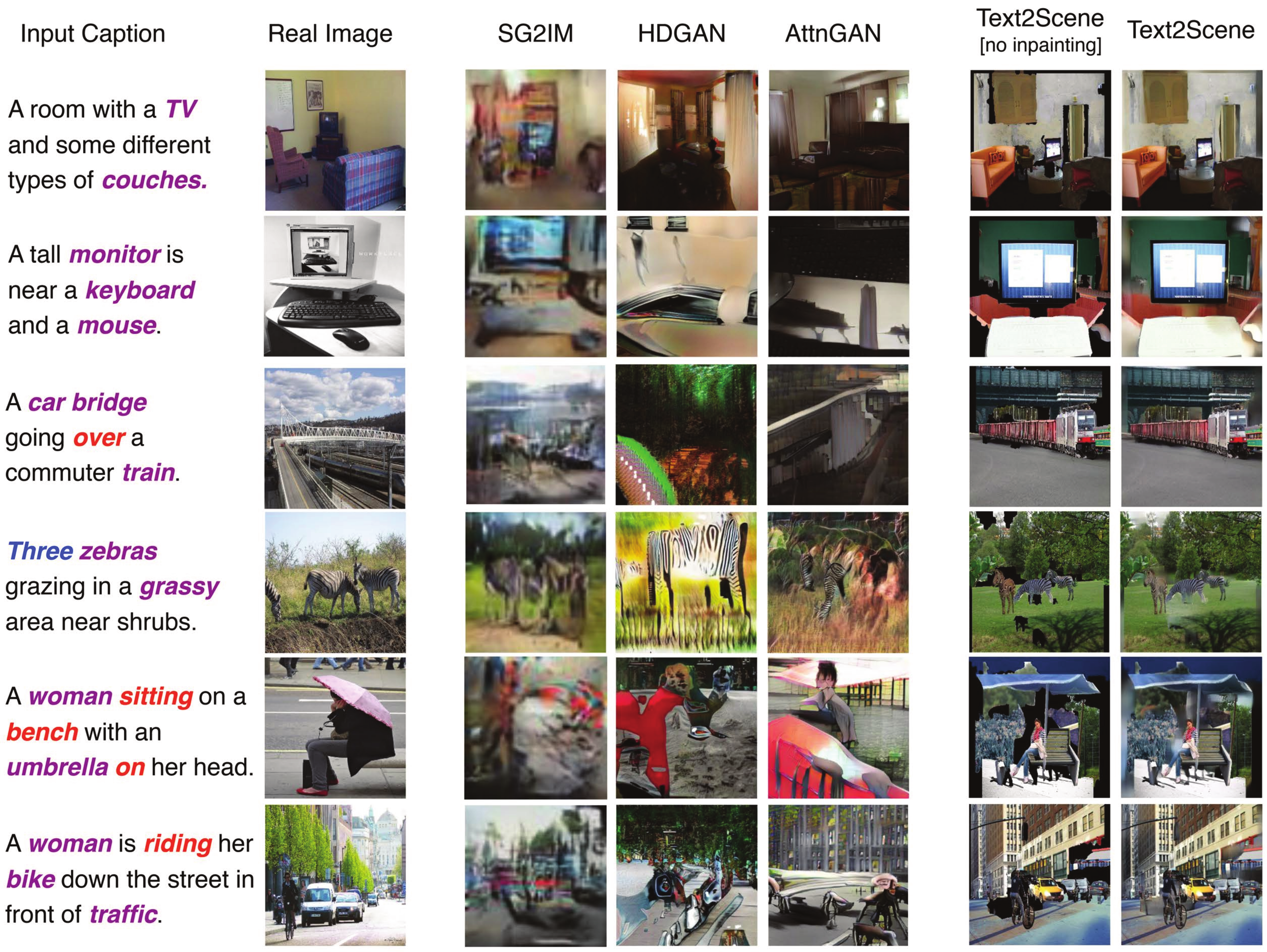}
  \caption{Qualitative examples of synthetic image generation (best viewed in color). The first column shows input captions with manually highlighted objects (purple), counts (blue) and relations (red). The second columns shows the \emph{true} images. Columns in the middle show competing approaches. The last two columns show the outputs of our model before and after pre-processing.}
  \label{fig:composite}
\end{figure*}

\noindent{\bf Abstract Scenes and Semantic Layouts:} Table \ref{quan_abstract} shows quantitative results on Abstract Scenes. 
Text2Scene improves over~\cite{Zitnick2013b} and our variants on all metrics except \texttt{U-obj Coord}. 
Human evaluation results on Table~\ref{human_abstract} confirm the quality of our outputs, where \texttt{Scores} are the percentage of sentence-scene pairs with a \texttt{true} entailment; ($\geq 1$) ($\geq 2$) indicate if our method produces scenes that entailed at least one (or two) out of three statements. Text2Scene also shows better results on statements with specific semantic information such as \texttt{Obj-single}, \texttt{Obj-pair}, and \texttt{Expression}, and is comparable on~\texttt{Location} statements. 
As a sanity check, we also test reference \emph{true} scenes provided in the Abstract Scenes dataset (first row). 
Results show that it is more challenging to generate the semantically related object pairs. 
Overall, the results also suggest that our proposed metrics correlate with human judgment on the task.

Figure \ref{fig:abstract_qual} shows qualitative examples of our models on Abstract Scenes in comparison with baseline approaches and the reference scenes. These examples illustrate that Text2Scene is able to capture semantic nuances such as the spatial relation between two objects (e.g., the bucket and the shovel are correctly placed in Jenny's hands in the last row) and object locations (e.g., Mike is on the ground near the swing set in the last row). 

Table \ref{quan_coco} shows an extrinsic evaluation on the layout generation task. We perform this evaluation by generating captions from our predicted layouts. Results show our full method generates the captions that are closest to the captions generated from true layouts. Qualitative results in Figure~\ref{fig:coco_qual} also show that our model learns important visual concepts such as presence and number of object instances, and their spatial relations.

\noindent{\bf Synthetic Image Composites:} Table~\ref{caption_coco} shows evaluation of synthetic scenes using automatic metrics. Text2Scene without any post-processing already outperforms all previous methods by large margins except AttnGAN~\cite{attngan}. 
As our model adopts a composite image generation framework without adversarial training, gaps between adjacent patches may result in unnaturally shaded areas. 
We observe that, after performing a regression-based inpainting~\cite{Semiparametric}, the composite outputs achieve consistent improvements on all automatic metrics. We posit that our model can be further improved by incorporating more robust post-processing or in combination with GAN-based methods. On the other hand, human evaluations show that our method significantly outperforms alternative approaches including AttnGAN~\cite{attngan}, demonstrating the potential of leveraging realistic image patches for text-to-image generation. It is important to note that SG2IM~\cite{sg2im} and Hong~et~al~\cite{txt2layout2img} also use segment and bounding box supervision -- as does our method, and AttnGan~\cite{attngan} uses an Imagenet (ILSVRC) pretrained Inceptionv3 network.
In addition, as our model contains a patch retrieval module, it is important that the model does not generate a synthetic image by simply retrieving patches from a single training image. 
On average, each composite image generated from our model contains 8.15 patches from 7.38 different source images, demonstrating that the model does not simply learn a global image retrieval.
Fig.~\ref{fig:composite} shows qualitative examples of the synthetic image composites, 
We include examples of generated images along with their corresponding source images from which patch segments are retrieved, and more extensive qualitative results in the supplemental material. Since our model learns about objects and relations separately, we also observed that it is often able to generalize to uncommon situations (as defined in~\cite{yatskar2017commonly}).
\section{Conclusions}
This work presents a novel sequence-to-sequence model for generating compositional scene representations from visually descriptive language.
We provide extensive quantitative and qualitative analysis of our model for different scene generation tasks on datasets from two different domains: Abstract Scenes~\cite{Zitnick2013b} and COCO~\cite{Lin2014}.
Experimental results demonstrate the capacity of our model to capture finer semantic concepts from visually descriptive text and generate complex scenes.

\vspace{0.05in}
\noindent{\bf Acknowledgements:} This work was partially supported by an IBM Faculty Award to V.O, and gift funding from SAP Research. \hfill

\appendix
\begin{figure}[ht!]
  \centering
  \Large\textbf{Supplementary Material}
\end{figure}
\section{Network Architecture}
Here we describe the network architectures for the components of our model in different tasks.

\subsection{Text Encoder}
We use the same network architecture for the text encoders in all our experiments, which consists of a single layer bidirectional recurrent network with Gated Recurrent Units (GRUs). 
It takes a linear embedding of each word as input and has a hidden dimension of 256 for each direction. 
We initialize the word embedding network with the pre-trained parameters from GloVe~\cite{glove}. 
The word embedding vectors are kept fixed for abstract scene and semantic layout generations but finetuned for synthetic image generation.

\subsection{Scene Encoder}
The scene encoder $\Omega$ for abstract scene generation is an Imagenet (ILSVRC) pre-trained ResNet-34~\cite{He2016}. 
Its parameters are fixed in all the experiments on Abstract Scene~\cite{Zitnick2013b}.
For layout and synthetic image generations, we develop our own scene encoders as the inputs for these tasks are not RGB images. 

Table \ref{table:layout_encoder} and \ref{table:volume_encoder} show the architecture details. 
Here $|\mathcal{V}|$ is the size of the categorical vocabulary. 
In the layout generation task, $|\mathcal{V}|$ is 83, including 80 object categories in COCO~\cite{Lin2014} and three special categorical tokens: $sos$, $eos$, $pad$, representing the start and end points for sequence generation and the padding token. 
For synthetic image generation, $|\mathcal{V}|$ is 98, including 80 object categories in COCO~\cite{Lin2014}, 15 supercategories for stuffs in COCO-stuff~\cite{cocostuff} and the special categorical tokens: $sos$, $eos$, $pad$. 

As described in the main paper, the input for synthetic image generation has a layer-wise structure where every three channels contain the color patches of a specific category from the background canvas image. 
In this case, the categorical information of the color patches can be easily learned.
On the other hand, since the input is a large but sparse volume with very few non-zero values, to reduce the number of parameters and memory usage, we use a depth-wise separable convolution as the first layer of $\Omega$ (index (2)), where each group of three channels (g3) is convolved to one single channel in the output feature map.

\begin{table}[t]
    \small\selectfont
    \centering
    \scalebox{0.95} {
        \begin{tabular}{|l|l|l|l|}
            \hline Index & Input & Operation & Output Shape \\
            \hline
            (1) &- & Input & $|\mathcal{V}|$ $\times$ 64 $\times$ 64\\
            (2) & (1) & Conv(7 $\times$ 7, $|\mathcal{V}|$ $\rightarrow$ 128, s2) & 128 $\times$ 32 $\times$ 32\\
            (3) & (2) & Residual(128 $\rightarrow$ 128, s1) & 128 $\times$ 32 $\times$ 32 \\
            (4) & (3) & Residual(128 $\rightarrow$ 256, s2) & 256 $\times$ 16 $\times$ 16 \\
            (5) & (4) & Bilateral upsampling & 256 $\times$ 28 $\times$ 28 \\
            \hline 
        \end{tabular}
    }
    \vspace{0.5mm}
    \caption{Architecture of our scene encoder $\Omega$ for layout generation. We follow the notation format used in~\cite{sg2im}. Here $|\mathcal{V}|$ is the size of the categorical vocabulary. The input and output of each layer have a shape of $C$ $\times$ $H$ $\times$ $W$, where $C$ is the number of channels and $H$ and $W$ are the height and width. The notation \textit{Conv}($K$ $\times$ $K$, $C_{in}$ $\rightarrow$ $C_{out}$) represents a convolutional layer with $K$ $\times$ $K$ kernels, $C_{in}$ input channels and $C_{out}$ output channels. The notation s2 means the convolutional layer has a stride of 2. The notation \textit{Residual}($C_{in}$ $\rightarrow$ $C_{out}$) is a residual module consisting of two 3 $\times$ 3 convolutions and a skip-connection layer. 
    In the first residual block (index (3)), the skip-connection is an identity function and the first convolution has a stride of 1 (s1). In the second residual block (index (4)),  the skip-connection is a 1 $\times$ 1 convolution with a stride of 2 (s2) and the first convolution also has a stride of 2 to downsample the feature map. Here all the convolutional layers are followed by a ReLU activation.}
    \label{table:layout_encoder}
\end{table}

\begin{table}[t]
    \small\selectfont
    \centering
    \scalebox{0.85} {
        \begin{tabular}{|l|l|l|l|}
            \hline Index & Input & Operation & Output Shape \\
            \hline
            (1) &- & Input & 3$|\mathcal{V}|$ $\times$ 128 $\times$ 128\\
            (2) & (1) & Conv(7 $\times$ 7, 3$|\mathcal{V}|$ $\rightarrow$ $|\mathcal{V}|$, s2, g3) & $|\mathcal{V}|$ $\times$ 64 $\times$ 64\\
            (3) & (2) & Residual($|\mathcal{V}|$ $\rightarrow$ $|\mathcal{V}|$, s1) & $|\mathcal{V}|$ $\times$ 64 $\times$ 64\\
            (4) & (3) & Residual($|\mathcal{V}|$ $\rightarrow$ 2$|\mathcal{V}|$, s1) & 2$|\mathcal{V}|$ $\times$ 64 $\times$ 64\\
            (5) & (4) & Residual(2$|\mathcal{V}|$ $\rightarrow$ 2$|\mathcal{V}|$, s1) & 2$|\mathcal{V}|$ $\times$ 64 $\times$ 64\\
            (6) & (5) & Residual(2$|\mathcal{V}|$ $\rightarrow$ 3$|\mathcal{V}|$, s2) & 3$|\mathcal{V}|$ $\times$ 32 $\times$ 32\\
            (7) & (6) & Residual(3$|\mathcal{V}|$ $\rightarrow$ 3$|\mathcal{V}|$, s1) & 3$|\mathcal{V}|$ $\times$ 32 $\times$ 32\\
            (8) & (7) & Residual(3$|\mathcal{V}|$ $\rightarrow$ 4$|\mathcal{V}|$, s1) & 4$|\mathcal{V}|$ $\times$ 32 $\times$ 32\\
            \hline 
        \end{tabular}
    }
    \vspace{0.05in}
    \caption{Architecture of our scene encoder $\Omega$ for synthetic image generation. The notations are in the same format of Table~\ref{table:layout_encoder}. The first convolution (index (2)) is a depthwise separable convolution where each group of three channels (g3) is convolved to one single channel in the output feature map. All the convolutional layers are followed by a LeakyReLU activation with a negative slope of 0.2.}
    \label{table:volume_encoder}
\end{table}

\subsection{Convolutional Recurrent Module}
The scene recurrent module for all our experiments is a convolutional GRU network~\cite{convgru} with one ConvGRU cell.
Each convolutional layer in this module have a 3 $\times$ 3 kernel with a stride of 1 and a hidden dimension of 512.
We pad the input of each convolution so that the output feature map has the same spatial resolution as the input.
The hidden state is initialized by spatially replicating the last hidden state from the text encoder.

\subsection{Object and Attribute Decoders}
Table~\ref{table:decoders} shows the architectures for our object and attribute decoders.
$\Psi^{o}$ and $\Psi^{a}$ are the spatial attention modules consisting of two convolutional layers. 
$\Theta^o$ is a two-layer perceptron predicting the likelihood of the next object using a softmax function. 
$\Theta^a$ is a four-layer CNN predicting the likelihoods of the location and attributes of the object.
As explained in the main paper, the output of $\Theta^a$ has $1+\sum_k|R^k|$ channels, where $|R^k|$ denotes the discretized range of the k-th attribute, or the dimension of the appearance vector $Q_t$ used as the query for patch retrieval for synthetic image generation. 
The first channel of the output from $\Theta^a$ predicts the location likelihoods which are normalized over the spatial domain using a softmax function. 
The rest channels predict the attributes for every grid location. 
During training, the likelihoods from the ground-truth locations are used to compute the loss. 
At each step of the test time, the top-1 location is first sampled from the model. The attributes are then collected from this sampled location. The text-based attention modules are defined similarly as in~\cite{attention}. When denoting $d_i = [h_{i}^E; x_i]$, $s^o_t = [u^o_t;o_{t-1}]$, and $s^a_t = o_{t}$, $\Phi^o$ and $\Phi^a$ are defined as:
\begin{eqnarray*} 
    c^*_t = \Phi^*(s^*_t, \{d_i\}) &=&  \sum_i \frac{\exp(\mathrm{score}(s^*_t, d_i))}{\sum_j\exp(\mathrm{score}(s^*_t, d_j))}\cdot d_i\\
    \mathrm{score}(s^*_t, d_k) &=& (s^*_t)^\intercal W_\Phi^*d_k, ~~~*\in{o, a}
\end{eqnarray*}

\noindent Here, $W_\Phi^o$ and $W_\Phi^a$ are trainable matrices which learn to compute the attention scores for collecting the context vectors $c^o_t$ and $c^a_t$.

These architecture designs are used for all the three generation tasks. 
The only difference is the grid resolution (H, W). 
For abstract scene and layout generations, (H, W) = (28, 28). 
For synthetic image generation, (H, W) = (32, 32).
Note that, although our model uses a fixed grid resolution, the composition can be performed on canvases of different sizes.

\begin{table}[t]
    \small\selectfont
    \centering
    \scalebox{0.7} {
        \begin{tabular}{|l|l|l|l|l|}
            \hline Module & Index & Input & Operation & Output Shape \\
            \hline 
            $\Psi^{o}$ & (1) & - & Conv(3$\times$3, 512$\rightarrow$256) & 256 $\times$ H $\times$ W \\
            &(2) &(1)& Conv(3$\times$3, 256$\rightarrow$1) & 1 $\times$ H $\times$ W \\
            \hline
            $\Psi^{a}$ & (1) & - & Conv(3$\times$3, 1324$\rightarrow$256) & 256 $\times H \times W$ \\
            &(2) & (1) & Conv(3$\times$3, 256$\rightarrow$1) & 1 $\times$ H $\times$ W \\
            \hline 
            $\Theta^o$ &(1)& - & Linear((1324 + $|\mathcal{V}|$)$\rightarrow$512) & 512\\
            &(2) & (1) & Linear(512$\rightarrow|\mathcal{V}|$) & $|\mathcal{V}|$ \\
            \hline 
            $\Theta^a$ &(1)& - 
            & Conv(3$\times$3, (1324+$|\mathcal{V}|$)$\rightarrow$512) & 512 $\times$ H $\times$ W\\
            &(2) & (1) & Conv(3$\times$3, 512$\rightarrow$256) & 256 $\times$ H $\times$ W \\
            &(3) & (2) & Conv(3$\times$3, 256$\rightarrow$256) & 256 $\times$ H $\times$ W \\
            &(4) & (3) & Conv(3$\times$3, 256$\rightarrow$($1+\sum_k|R^k|$)) & ($1+\sum_k|R^k|$) $\times$ H $\times$ W  \\
            \hline 
        \end{tabular}
    }
    \vspace{0.5mm}
    \caption{Architectures for the object and attribute decoders. The notation \textit{Linear}($C_{in}$ $\rightarrow$ $C_{out}$) represents a fully connected layer with $C_{in}$ input channels and $C_{out}$ output channels. All layers, except the last layer of each module, are followed by a ReLU activation.}
    \label{table:decoders}
\end{table}

\subsection{Foreground Patch Embedding}
The foreground segment representation we use is similar with the one in~\cite{Semiparametric}, where each segment $P$ is represented by a tuple ($P^{color}$, $P^{mask}$, $P^{context}$).
Here $P^{color} \in \mathbb{R}^{3 \times H \times W}$ is a color patch containing the segment, $P^{mask} \in \{0, 1\}^{1 \times H \times W}$ is a binary mask indicating the foreground region of $P^{color}$, $P^{context} \in \{0, 1\}^{|\mathcal{V}| \times H \times W} $ is a semantic map representing the
semantic context around $P$.  
The context region of $P$ is obtained by computing the bounding box of the segment and enlarging it by 50\% in each direction. 

Table~\ref{table:patch_encoder} shows the architecture of our foreground patch embedding network. 
Here, the concatenation of ($P^{color}$, $P^{mask}$, $P^{context}$) is fed into a five-layer convolutional network which reduces the input into a 1D feature vector $F_s$ (index (7)). 
As this convolutional backbone is relatively shallow, $F_s$ is expected to encode the shape, appearance, and context, but may not capture the fine-grained semantic attributes of $P$. 
In our experiments, we find that incorporating the knowledge from the pre-trained deep features of $P^{color}$ can help retrieve segments associated with strong semantics, such as the "person" segments. 
Therefore, we also use the pre-trained features $F_d$ (index (8)) of $P^{color}$ from the mean pooling layer of ResNet152~\cite{He2016}, which has 2048 features.
The final vector $F_t$ is predicted from the concatenation of ($F_s$, $F_d$) by a linear regression.

\begin{table}[t]
    \small\selectfont
    \centering
    \scalebox{0.85} {
        \begin{tabular}{|l|l|l|l|}
            \hline Index & Input & Operation & Output Shape \\
            \hline
            (1) &- & Input layout & ($|\mathcal{V}|$ + 4) $\times$ 64 $\times$ 64\\
            (2) & (1) & Conv(2 $\times$ 2, ($|\mathcal{V}|$ + 4) $\rightarrow$ 256, s2) & 256 $\times$ 32 $\times$ 32\\
            (3) & (2) & Conv(2 $\times$ 2, 256 $\rightarrow$ 256, s2) & 256 $\times$ 16 $\times$ 16\\
            (4) & (3) & Conv(2 $\times$ 2, 256 $\rightarrow$ 256, s2) & 256 $\times$ 8 $\times$ 8\\
            (5) & (4) & Conv(2 $\times$ 2, 256 $\rightarrow$ 256, s2) & 256 $\times$ 4 $\times$ 4\\
            (6) & (5) & Conv(2 $\times$ 2, 256 $\rightarrow$ 128, s2) & 256 $\times$ 2 $\times$ 2\\
            (7) & (6) & Global average pooling & 256 \\
            \hline
            (8) &- & Input patch feature & 2048\\
            (9) & (7)(8) & Linear((256 + 2048) $\rightarrow$ 128) & 128\\
            \hline 
        \end{tabular}
    }
    \vspace{0.5mm}
    \caption{Architecture of our foreground patch embedding network for synthetic image generation. All the convolutional layers are followed by a LeakyReLU activation with a negative slope of 0.2.}
    \label{table:patch_encoder}
\end{table}

\subsection{Inpainting Network}
Our inpainting network has the same architecture as the image synthesis module proposed in~\cite{Semiparametric}, except that we exclude all the layer-normalization layers. 
To generate the simulated canvases on COCO, we follow the procedures proposed in~\cite{Semiparametric}, but make minor modifications: (1) we use the trained embedding patch features to retrieve alternative segments to stencil the canvas, instead of the intersection-over-union based criterion used in~\cite{Semiparametric}. (2) we do not perform boundary elision for the segments as it may remove fine grained details of the segments such as human faces.

\section{Optimization}
For optimization we use Adam~\cite{Diederik2015} with an initial learning rate of $5e-5$. 
The learning rate is decayed by $0.8$ every $3$ epochs.
We clip the gradients in the back-propagation such that the norm of the gradients is not larger than 10.
Models are trained until validation errors stop decreasing. 
For abstract scene generation, we set the hyperparameters ($w_o$, $w_l$, $w_{pose}$, $w_{expression}$, $w_{size}$, $w_{direction}$, $w_a^O$, $w_a^A$) to (8,2,2,2,1,1,1,1).
For semantic layout generation, we set the hyperparameters ($w_o$, $w_l$, $w_{size}$, $w_{aratio}$, $w_a^O$, $w_a^A$) to (5,2,2,2,1,0).
For synthetic image generation, we set the hyperparameters ($w_o$, $w_l$, $w_{size}$, $w_{aratio}$, $w_a^O$, $w_a^A$, $w_e$, $\alpha$) to (5,2,2,2,1,0,10,0.5). 
The hyperparameters are chosen to make the losses of different components comparable.
Exploration of the best hyperparameters is left for future work.

\begin{figure}[t]
  \centering
  \includegraphics[width=0.4\textwidth]{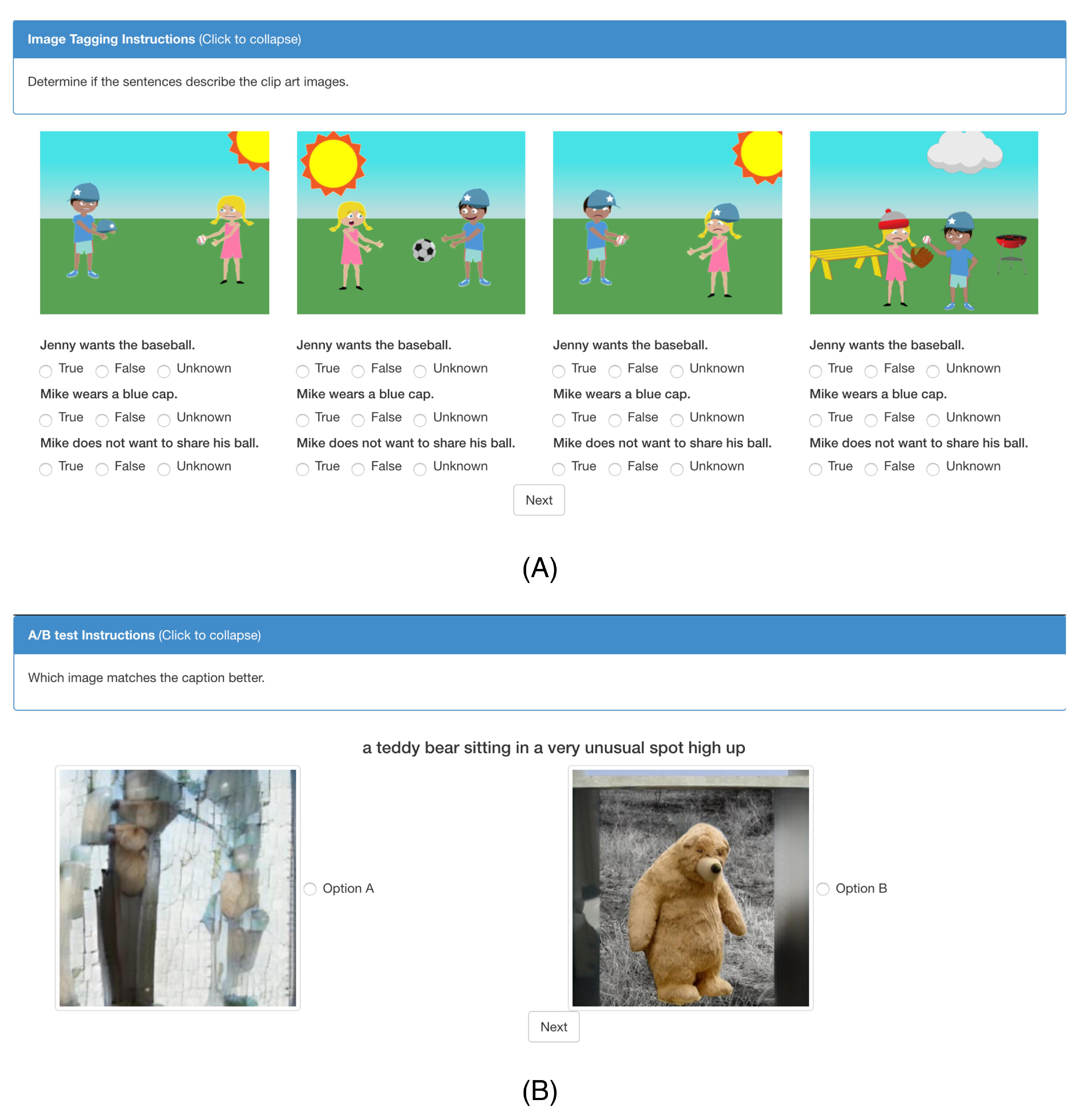}
  \caption{Screen shots of the user interfaces for our human subject studies on Amazon Mechanical Turk. (A) User interface for the evaluation study of the abstract scene generation experiment; (B) User interface for the evaluation study of the synthetic image generation experiment.}
  \label{fig:user_study}
\end{figure}

\vspace{-0.01in}
\section{User Study}
We conduct two user studies on Amazon Mechanical Turk (AMT).

The first user study is to evaluate if the generated clip-art scenes match the input sentences. 
To this end, we randomly select 100 groups of images generated from the sentences in the test set.
Each group consists of three images generated by different models, and the ground truth reference image.
During the study, these images and the corresponding sentences are presented in random orders.
The human annotators are asked to determine if the entailment between the generated scene and the sentence is \texttt{true}, \texttt{false} or \texttt{uncertain}.
Each group of images is seen by three annotators. 
We ignore the \texttt{uncertain} responses and report the results using majority opinions. Figure~\ref{fig:user_study} (A) shows the user interface of this study.

The second user study is on the synthetic image generation task, where we compare the generated images from our model and three state-of-the-art approaches: SG2IM~\cite{sg2im}, HDGAN~\cite{hdgan}, and AttnGAN~\cite{attngan}.
In each round of the study, the human annotator is presented with one sentence and two generated images: one from our model, the other from an alternative approach.
The orders of the images are randomized.
We ask the human annotator to select the image which matches the sentence better.
In total, we collect results for 500 sentences randomly selected from the test set, using five annotators for each. 
Figure~\ref{fig:user_study} (B) shows the user interface of this study.

\begin{figure*}[t]
  \centering
  \includegraphics[width=0.98\textwidth]{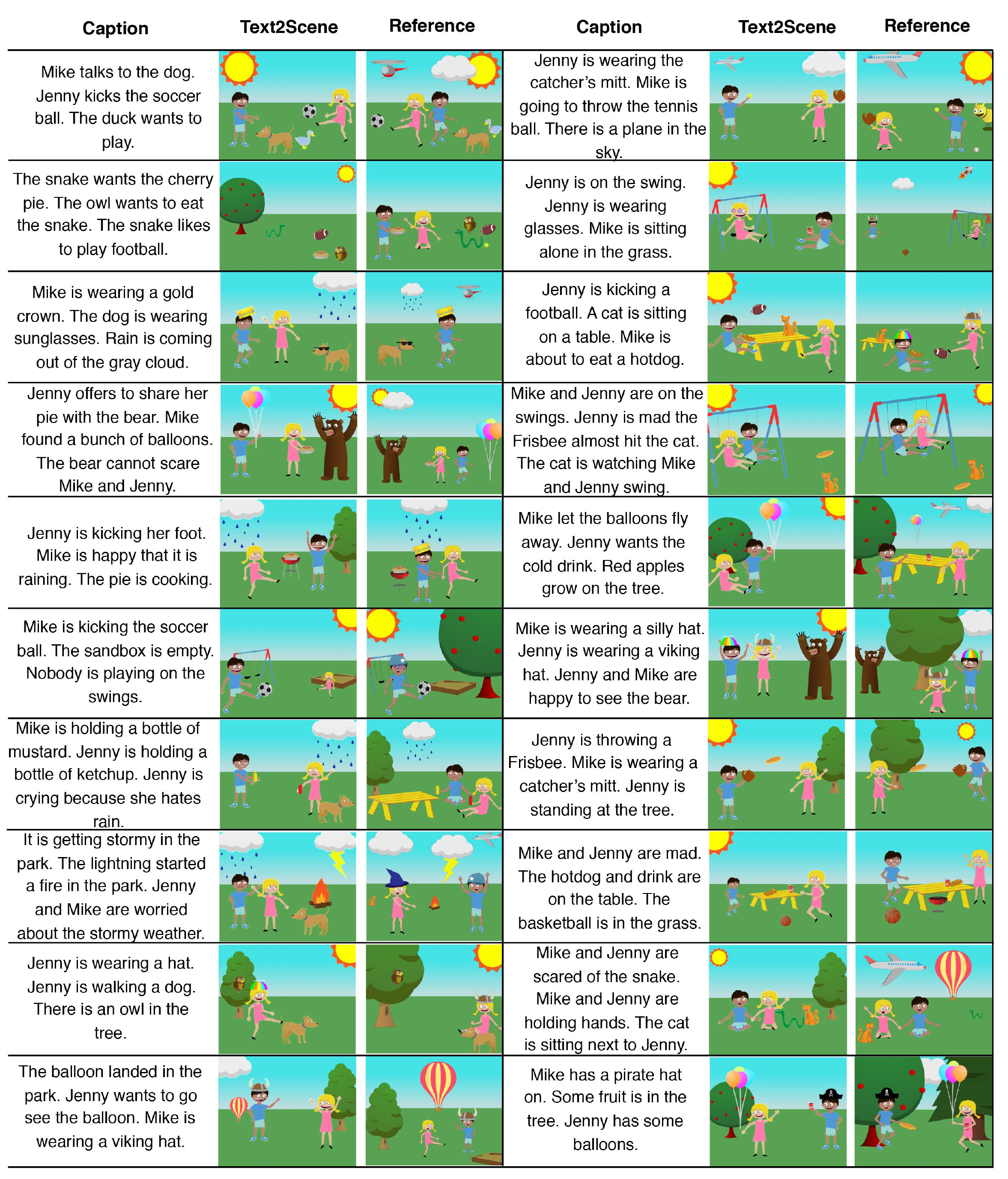}
  \caption{More qualitative examples of the abstract scene generation experiment.}
  \label{fig:abstract_sup}
\end{figure*}

\section{More qualitative examples}
\vspace{-0.01in}
\subsection{Abstract Scene}
We present more qualitative examples on Abstract Scene~\cite{Zitnick2013b} in Fig.~\ref{fig:abstract_sup}. The examples show that our model does not simply replicate the ground truth reference scenes, but generates dramatically different clip-arts which still match the input textual descriptions.
\vspace{-0.01in}
\subsection{Layout Generation}
\begin{figure*}[t]
  \centering
  \includegraphics[width=0.95\textwidth]{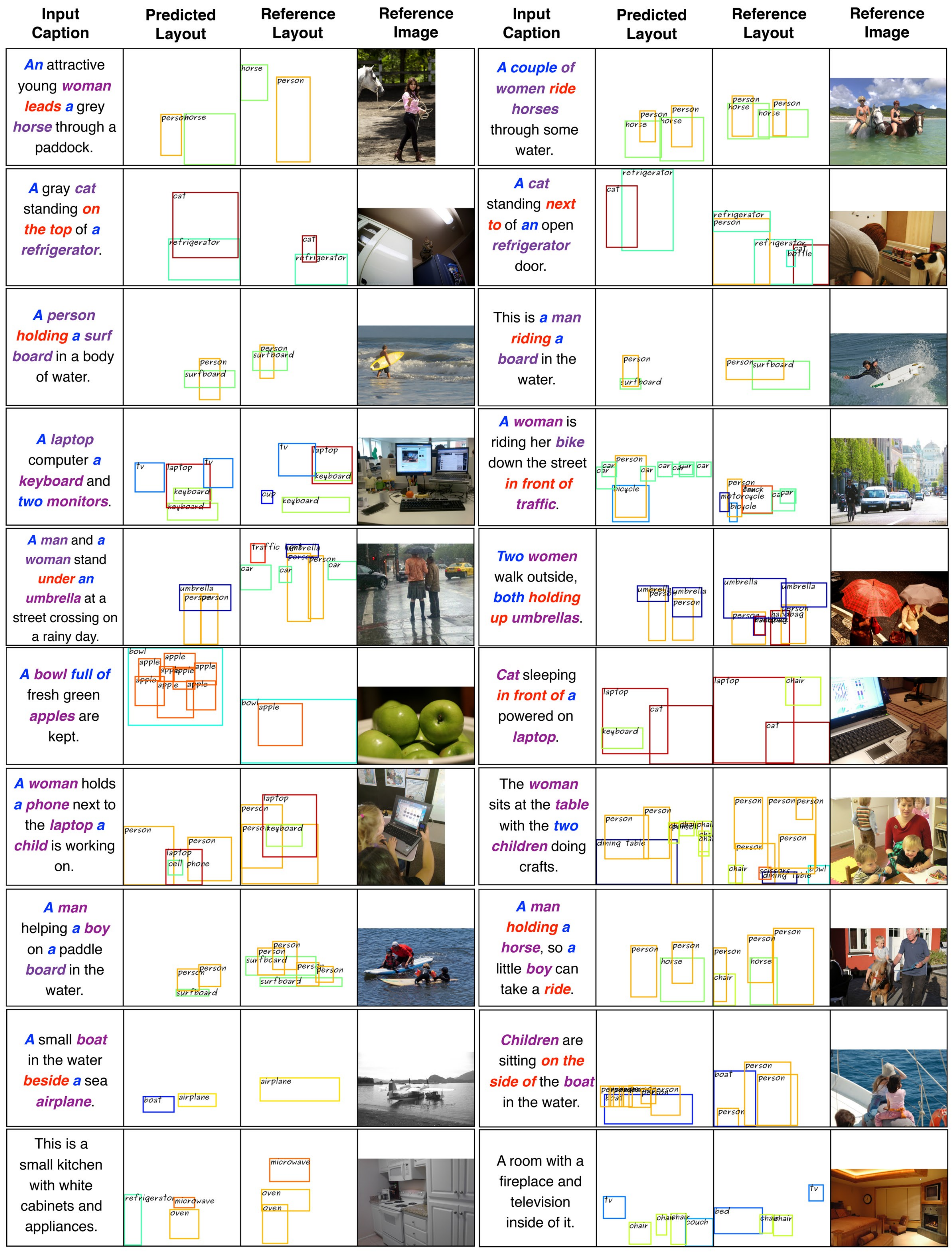}
  \caption{More qualitative examples of the layout generation experiment (best viewed in color). The presences (purple), counts (blue), and spatial relations (red) of the objects are highlighted in the captions. The last row shows the cases when the layouts are underspecified in the input captions.}
  \label{fig:layout_sup}
\end{figure*}

We present more qualitative examples for layout generation in Fig.~\ref{fig:layout_sup}. 
The examples include various scenes containing different object categories. 
Our model manages to learn important semantic concepts from the language, such as the presence and count of the objects, and their spatial relations.
\vspace{-0.01in}
\subsection{Synthetic Image Generation}
\begin{figure*}[t]
  \centering
  \includegraphics[width=0.9\textwidth]{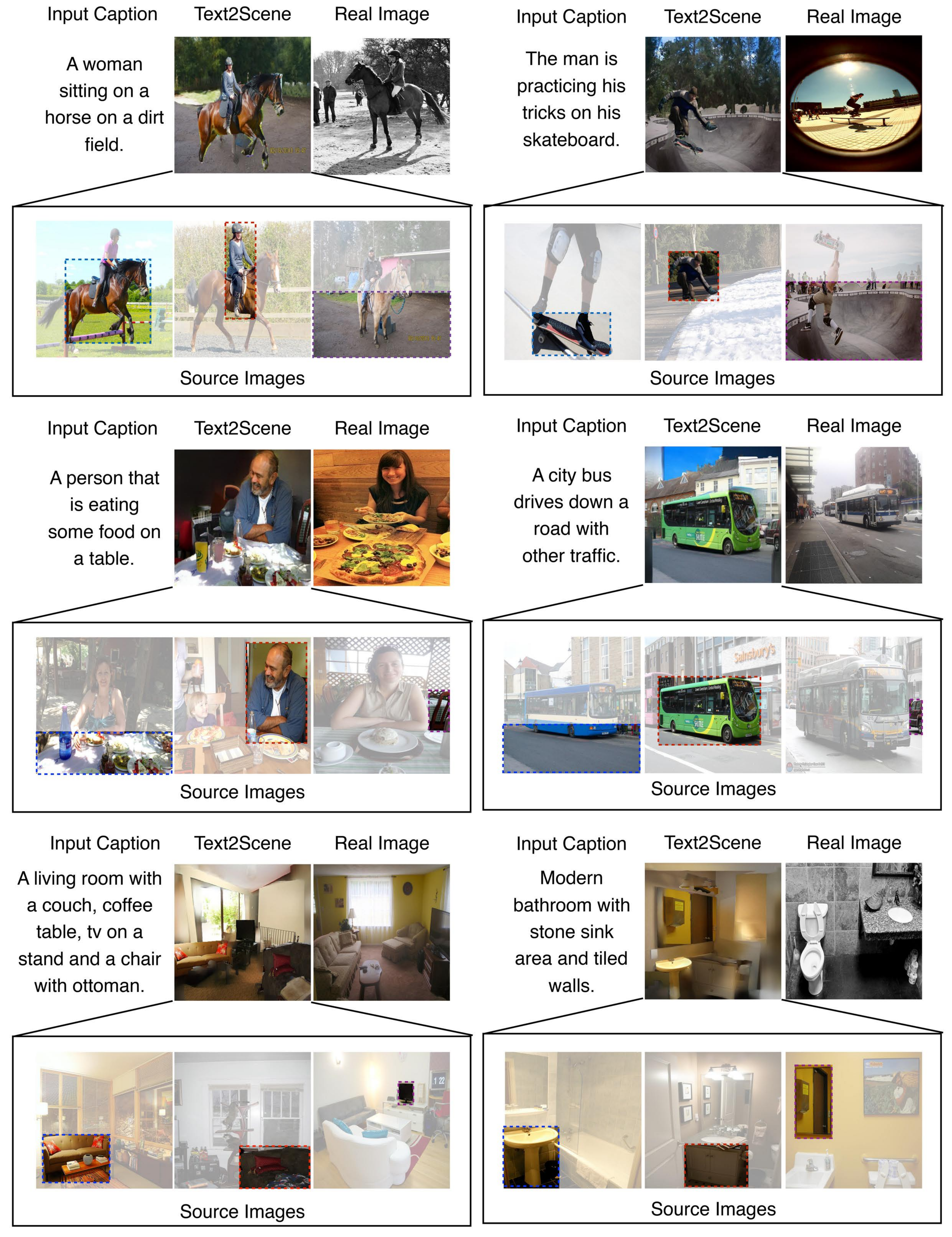}
  \caption{Example synthetic images and the source images from which the patch segments are retrieved for compositing. For each synthetic image, we show three source images for clarity.}
  \label{fig:comp_src}
\end{figure*}

\begin{figure*}[t]
  \centering
  \includegraphics[width=0.9\textwidth]{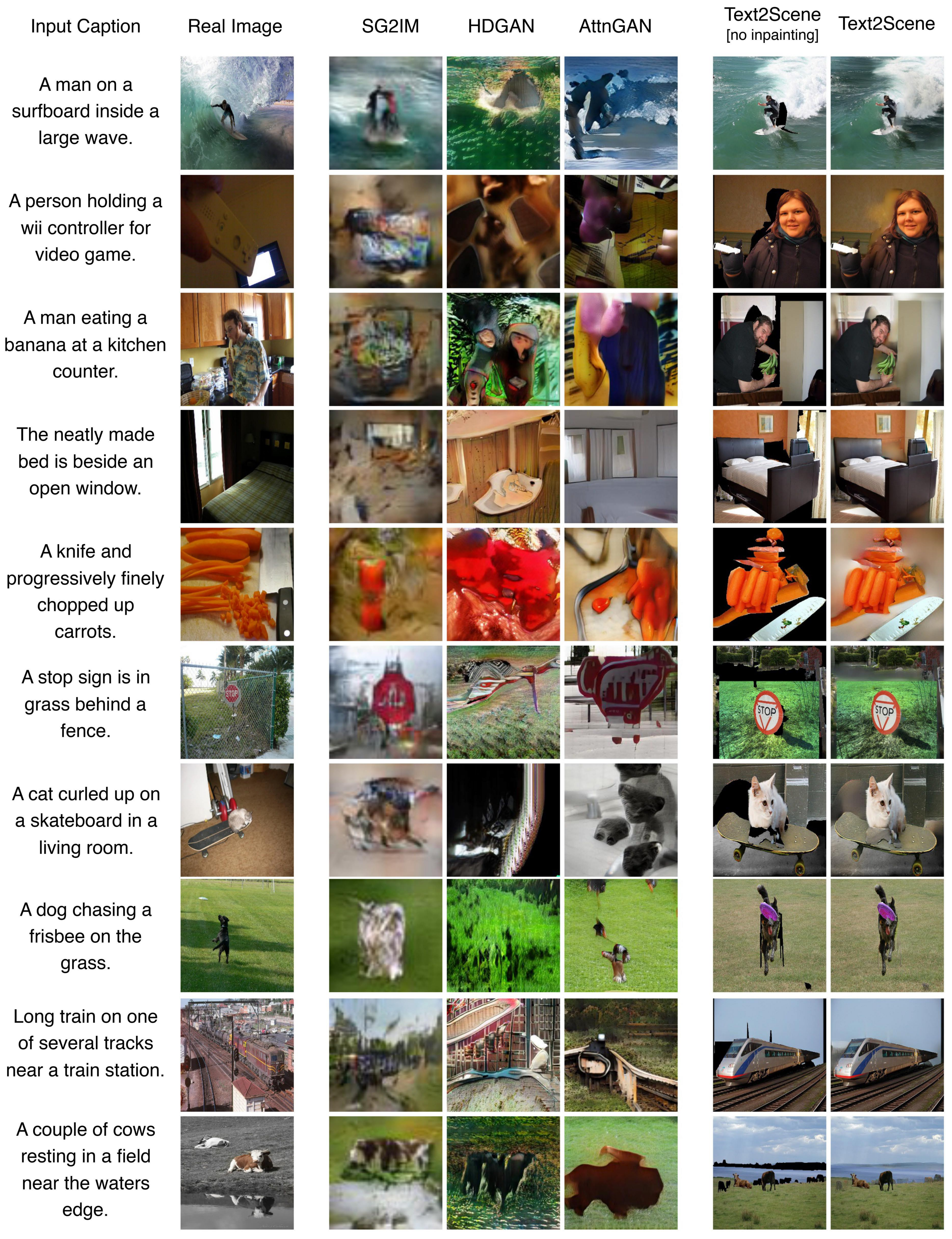}
  \caption{More qualitative examples of the synthetic image generation experiment. }
  \label{fig:composite_sup}
\end{figure*}

To demonstrate our model does not learn an image-level retrieval on the training set, we present in Fig.~\ref{fig:comp_src} the generated images and the corresponding source images from which the patch segments are retrieved for compositing.
For each generated image, we show three source images for clarity.
The examples illustrate that our model learns not only the presence and spatial layout of objects, but also the semantic knowledge that helps retrieve segments in similar contexts. Fig.~\ref{fig:composite_sup} shows more qualitative examples of our model for synthetic image generation.

\clearpage

\clearpage

\clearpage

\clearpage
\clearpage

{\small
\bibliographystyle{ieee}
\bibliography{egbib}
}

\end{document}